\documentclass[conference]{IEEEtran}
\IEEEoverridecommandlockouts
\usepackage{cite}
\usepackage{amsmath,amssymb,amsfonts}
\usepackage{algorithmic}
\usepackage{graphicx}
\usepackage{multirow}
\usepackage{textcomp}
\usepackage[table]{xcolor}
\usepackage{bm}
\usepackage{array}
\definecolor{lightred}{RGB}{255, 230, 230} % light pinkish red

\def\BibTeX{{\rm B\kern-.05em{\sc i\kern-.025em b}\kern-.08em
    T\kern-.1667em\lower.7ex\hbox{E}\kern-.125emX}}

\usepackage{etoolbox}

\usepackage{caption} 

\newcommand{\insertfig}{\includegraphics[width=0.75\linewidth]{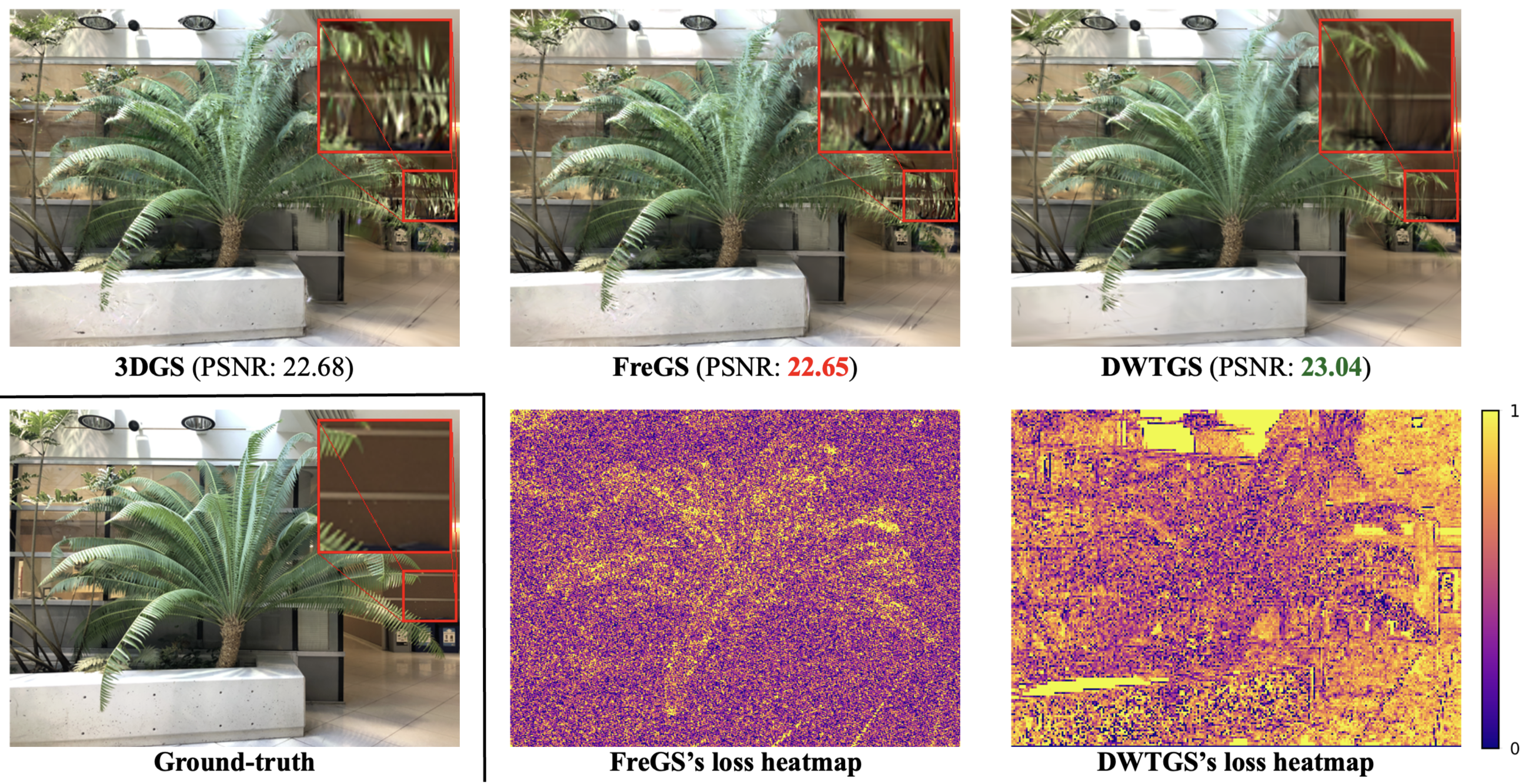}\captionof{figure}{We propose DWTGS, a framework that performs joint spatial-frequency regularization for sparse-view 3DGS \cite{3DGS}. DWTGS (right) introduces a wavelet-space loss function that primarily supervises low-frequency (LF) details, as indicated by the brighter colors at relatively homogeneous regions in the per-pixel gradient heatmap of the loss. This effectively reduces high-frequency (HF) hallucinations, as compared to vanilla 3DGS (left) and a closely related work, FreGS \cite{FreGS} (middle).}}\label{fig_teaser_dwtgs}\addtocounter{figure}{-2}

\makeatletter
\apptocmd{\@maketitle}{\centering\insertfig}{}{}% insert the figure after authors
\makeatother

\begin{document}

\title{DWTGS: Rethinking Frequency Regularization for Sparse-view 3D Gaussian Splatting

}

\author{
\IEEEauthorblockN{Hung Nguyen, Runfa Li, An Le, Truong Nguyen}
\IEEEauthorblockA{
\textit{Video Processing Lab, Department of Electrical \& Computer Engineering} \\
\textit{University of California San Diego} \\
\{hun004, rul002, d0le, tqn001\}@ucsd.edu}
}

% \and
% \and
% \and
% \and
% \and

\maketitle

\begin{abstract}
Sparse-view 3D Gaussian Splatting (3DGS) presents significant challenges in reconstructing high-quality novel views, as it often overfits to the widely-varying high-frequency (HF) details of the sparse training views. While frequency regularization can be a promising approach, its typical reliance on Fourier transforms causes difficult parameter tuning and biases towards detrimental HF learning. We propose DWTGS, a framework that rethinks frequency regularization by leveraging wavelet-space losses that provide additional spatial supervision. Specifically, we supervise only the low-frequency (LF) LL subbands at multiple DWT levels, while enforcing sparsity on the HF HH subband in a self-supervised manner. Experiments across benchmarks show that DWTGS consistently outperforms Fourier-based counterparts, as this LF-centric strategy improves generalization and reduces HF hallucinations.
\end{abstract}


\section{Introduction}

3D Gaussian Splatting (3DGS) \cite{3DGS} has emerged as a state-of-the-art method for reconstructing 3D scenes from 2D images, thus generating high-quality, photorealistic novel views. By modeling scenes with Gaussian primitives, it significantly accelerates training compared to earlier methods like NeRF \cite{NeRF} and has been integrated into various practical applications \cite{DynaGSSLam, SplatSDF, MonoSelfRecon} that necessitate 3D reconstruction and understanding. However, 3DGS typically relies on dense training views with precise camera poses, demanding extensive and accurate data collection. Without dense views, the reconstructed 3D geometry is inadequately constrained, causing scene explosions or artifacts that severely compromise rendering quality. This restricts its applicability in real-world scenarios, where such data might be unfeasible to obtain \cite{SparseGS}. 

Therefore, sparse-view 3DGS becomes an important and actively explored problem. Multiple authors leverage prior information, e.g., depth \cite{depth-regularized3DGS, depth-aware3DGS, dngaussian}, feature matching consistency \cite{SCGaussian, FewViewGS}, to regularize 3DGS training or to improve the Gaussian initialization. Alternatively, diffusion models have been introduced to synthesize pseudo training views, thus emulating dense-view training \cite{RI3D, 3DGS-Enhancer, Deceptive3DGS}. However, those methods usually lack adaptation to the training images, or incur prohibitive pre-training time.

FreeNeRF \cite{FreeNeRF} first demonstrates the surprising effectiveness of prior-free frequency regularization. It demonstrates that sparse‑view NeRF suffers from HF overfitting, as simply suppressing HF inputs and relying only on LF in early training markedly improves results. Building on this idea at the loss‑function level, where supervision is decoupled from the specific inputs (e.g., 3D coordinates, Gaussians), FreGS \cite{FreGS} and PGDGS \cite{PGDGS} utilize fully supervised, Fourier‑space losses that split LF and HF components, injecting the HF term only after an initial LF‑focused phase. However, because learning slowly-varying LF structures is relatively simple, this supervision ends up being biased towards HF details (middle column, Fig. 1), reverting to the HF overfitting weakness identified by FreeNeRF.

In this paper, we propose DWTGS, a loss function-based framework that instead regularizes frequency in the wavelet space. To enforce LF-centric supervision that mitigates HF overfitting, we leverage a multi-level loss on the ground-truth and render LL subbands. We discard HF consistency in favor of a self-supervised loss that encourages sparsity on the HF HH subband at novel views. Those designs improve rendering quality in our experiments, confirming that sparse-view 3DGS benefits more from LF supervision, similar to tasks like low-light enhancement which also do not have reliable HF information for supervision \cite{CutFreq}. Furthermore, moving to the wavelet space allows readily available frequency disentanglement, in contrast to Fourier space which additionally requires defining parameters for the low-pass and high-pass filtering operations \cite{imageprocessing_book}. In summary, our contributions are as follows:
\begin{itemize}
    \item We propose the DWTGS framework, leveraging LF-centric losses in wavelet space that only supervise the LL subband while encourage the sparsity of the HH subband to mitigate HF overfitting. 
    \item Through comprehensive experiments, we show that our DWTGS consistently outperforms Fourier counterparts in frequency regularization, and that sparse-view 3DGS benefits from informative, LF-centric approaches.  
\end{itemize}

\section{Related Works}

\textbf{Discrete Wavelet Transform (DWT) for Neural Rendering}. Recently, the DWT has gained increasing attraction in deep computer vision frameworks as it enables disentangled frequency learning \cite{WF-VAE}, multi-scale understanding \cite{WaveletConvolution} and flexible representation learning \cite{Orthogonal-LatticeUwU-app, BiorthogonalLifting}. While still limited, extensions of the concept to 3DGS are also being explored, e.g., for fine detail enhancement \cite{MicroMacro, 3D-GSW} and coarse-to-fine learning \cite{AutoOpti3DGS}. Within this line of works, our DWTGS framework novelly introduces the DWT to sparse-view 3DGS at the loss function level, representing an effective, spatially-aware method of frequency regularization.

\textbf{Frequency Regularization for Neural Rendering}. Pixel-wise losses such as $\mathcal{L}_1$ or $\mathcal{L}_2$ often exhibit low-frequency (LF) bias, impairing high-frequency (HF) detail reconstruction \cite{lowfreqbias_loss, WaveletLossesGenerative}. To address this, FreGS \cite{FreGS} introduces auxiliary HF supervision in the Fourier domain, mitigating over-smoothing artifacts in dense-view 3DGS. However, its effectiveness under sparse views remains unclear. While PGDGS \cite{PGDGS} adapts FreGS to this setting, it still operates in Fourier space, requiring extensive tuning of loss weights while being biased towards HF learning, which is prone to overfitting \cite{FreeNeRF}. Instead, our DWTGS leverages wavelet-space, LF-centric losses that effectively mitigate HF overfitting in sparse-view 3DGS.

% Furthermore, inspired by \cite{CutFreq}, which shows that different imaging problems might benefit differently from LF or HF supervision, we identify sparse-view 3DGS as LF-centric problem that benefits more from LF supervision. This makes intuitive sense as the sparse views available are inadequate to constrain the highly varying HF details, thus making HF supervision unreliable. In designing the wavelet losses, this observation motivates us to supervise only the LF, while merely encouraging sparsity for the HF.

% As will be shown in Section~\ref{subsec_struggle}, it struggles to represent high-frequency details, especially under sparse-view settings where they are under-constrained and vary significantly across scenes.

\section{Preliminary Background} \label{sec_prelim}

\subsection{3D Gaussian Splatting} \label{sec_prelim_3dgs}

Given multi-view 2D images of a scene, 3DGS~\cite{3DGS} reconstructs the scene in 3D using 3D Gaussian primitives, enabling photorealistic rendering from novel viewpoints. Each Gaussian is defined by optimizable parameters: center positions $\bm{\mu}$, opacities $\sigma$, covariances $\boldsymbol{\Sigma}$, and colors $\mathbf{c}$. The optimization is guided by a differentiable loss function:
\begin{equation}
\mathcal{L}_{3DGS} = (1-\lambda) \mathcal{L}_1(\mathbf{X}^{gt}, \mathbf{X}) + \lambda \mathcal{L}_{D\text{-}SSIM}(\mathbf{X}^{gt}, \mathbf{X})
\label{eq_3dgs_loss}
\end{equation}
where $\mathbf{X}^{gt}$ and $\mathbf{X}$ denote the ground-truth and render images from the same training viewpoint, respectively. $\mathcal{L}_1$ is the MAE loss, and $\mathcal{L}_{D\text{-}SSIM}$ supervises perceptual similarity \cite{SSIM}, with a balancing term $\lambda$. To adaptively control the Gaussians' densities, 3DGS proposes the ADC scheme \cite{3DGS}, which can remedy ``over-reconstruction'', which happens when high-variance 3D regions are covered only by a few coarse-sized Gaussians, leading to unseemly artifacts \cite{FreGS}. 

\subsection{Discrete Wavelet Transform (DWT)} \label{subsec_prelim_dwt}

\begin{figure}[t]
    \centering \includegraphics[width=\linewidth]{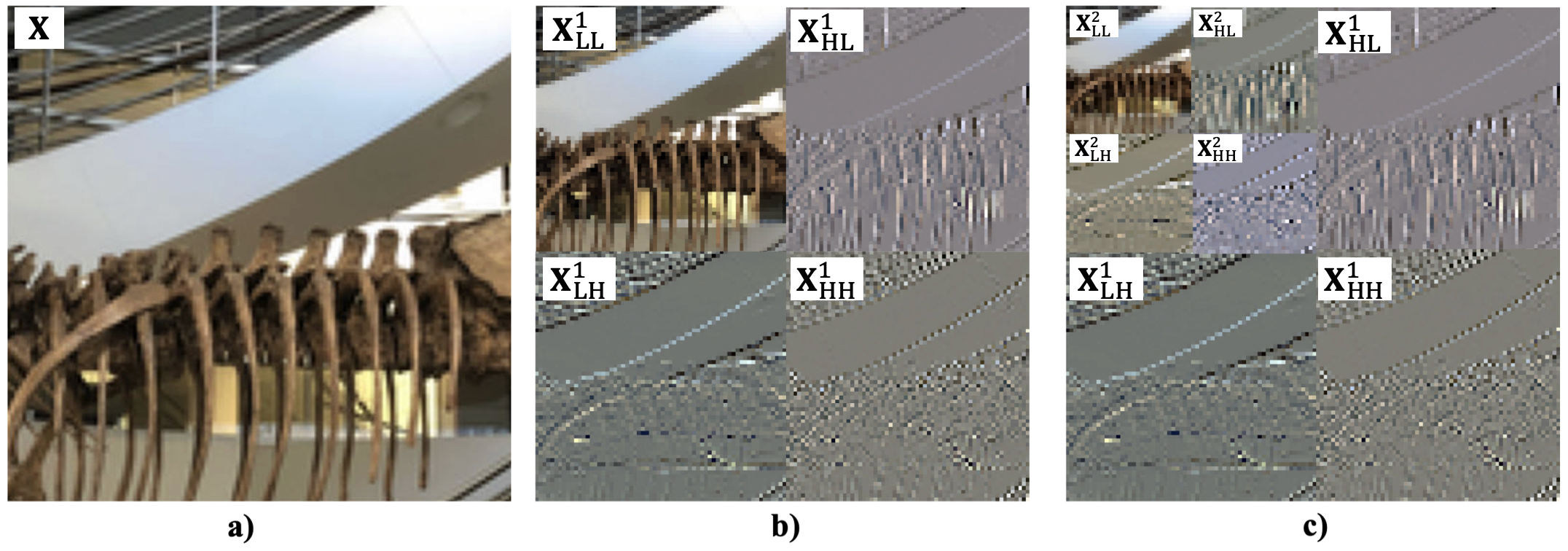}
    \caption{1-level (b) and 2-level DWT subbands (c) of an image region (a) \cite{LLFF}. Each subband provides directional frequency information, and gets coarser with increasing DWT level.}
    \label{fig_vis_subband_multi}
\end{figure}

\begin{figure*}[t]
    \centering
    \includegraphics[width=0.9\linewidth]{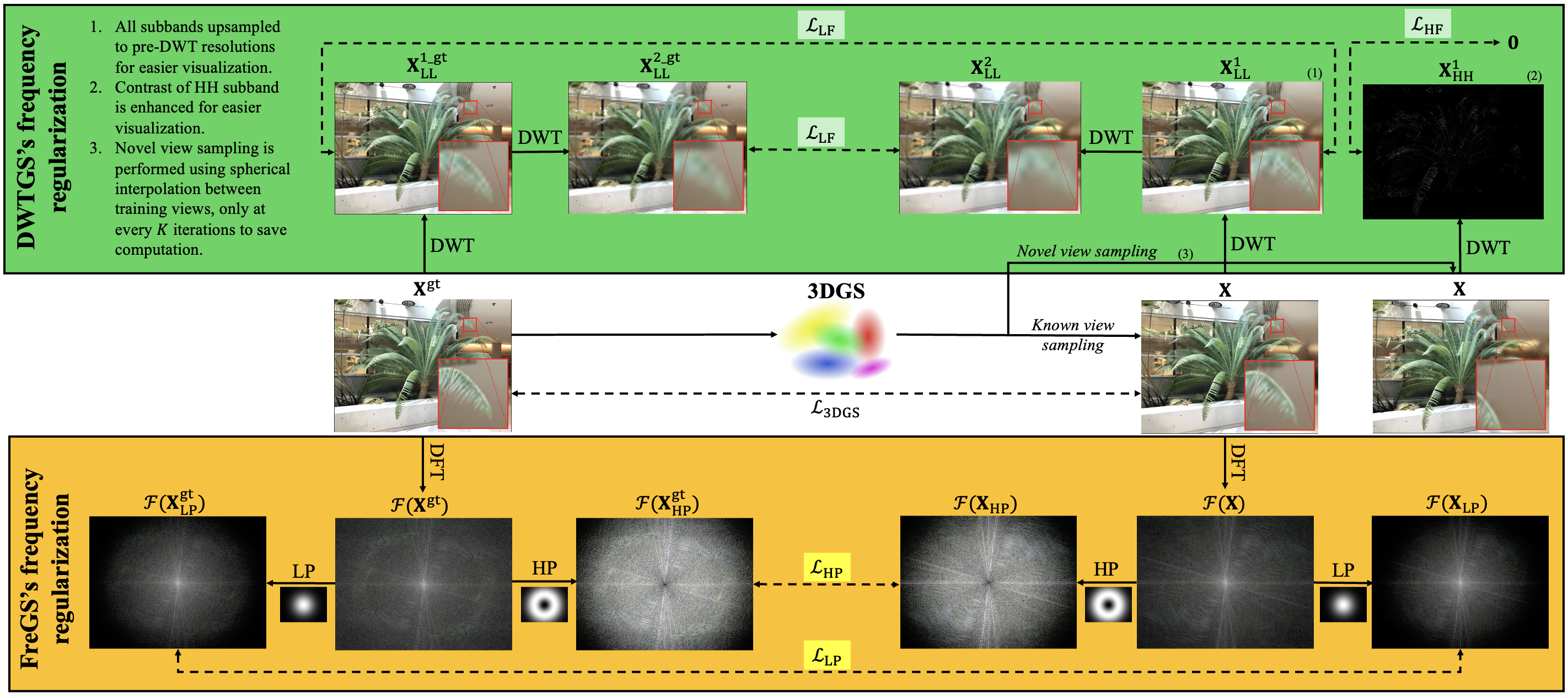}
    \caption{Architectures of our DWTGS framework (top) and FreGS \cite{FreGS} (bottom). Contrary to the latter's Fourier-space loss, DWTGS proposes wavelet-space losses for sparse-view 3DGS, which consist of a LF and HF subloss. The LF loss supervises multi-level GT and render LL subbands, while the HF loss enforces sparsity of the HH subband in novel views.}
    \label{fig_arch_dwtgs}
\end{figure*}

The Discrete Fourier Transform (DFT) excels in frequency analysis of digital images, providing crucial information about how frequency components contribute to the original image. However, based on a global exponential basis, it lacks spatial localization and interpretability \cite{imageprocessing_book}. On the other hand, the DWT simultaneously captures both frequency and spatial information. Given a 2D image $\mathbf{X}$, it provides the four  following subbands:
\begin{equation} \label{eq_fdwt}
\begin{split}
\mathbf{X}_{LL} = \mathbf{L}_0\mathbf{X}\mathbf{L}_1, \quad
\mathbf{X}_{LH} = \mathbf{H}_0\mathbf{X}\mathbf{L}_1, \\
\mathbf{X}_{HL} = \mathbf{L}_0\mathbf{X}\mathbf{H}_1, \quad
\mathbf{X}_{HH} = \mathbf{H}_0\mathbf{X}\mathbf{H}_1
\end{split}
\end{equation}
where $\mathbf{L}_{(\cdot)}$ and $\mathbf{H}_{(\cdot)}$ are the low-pass and high-pass matrices that filter the columns or rows of $\mathbf{X}$, respectively. The subscript $\{0, 1\}$ denotes filtering along columns or rows. Each filtering matrix is constructed by shifting columns or rows of a common 1D wavelet filter \cite{wavelet-book}. The LL subband is obtained by applying low-pass filters along both rows and columns, preserving the image’s coarse structure. The LH and HL subbands result from applying a high-pass filter in one direction and a low-pass filter in the other, highlighting horizontal and vertical details, respectively. Lastly, the HH subband is obtained by high-pass filtering in both directions, capturing fine diagonal details. By repeating \eqref{eq_fdwt} for the resulting LL subband, the multi-level DWT can be achieved, as shown in Fig.~\ref{fig_vis_subband_multi}.

\section{Methodology}

\subsection{Fourier-space Frequency Regularization}

FreGS \cite{FreGS} shows that the densification scheme introduced in Section~\ref{sec_prelim_3dgs} is inadequate to remedy the over-reconstruction issue and represent high-frequency (HF) details, as the pixel-wise supervisions in \eqref{eq_3dgs_loss} have a low-frequency (LF) bias \cite{lowfreqbias_loss}. Therefore, to disentangle frequency learning, it proposes an auxiliary loss that supervises between the ground-truth (GT) and render frequency spectra with magnitude $|(\cdot)|$ and phase $\angle(\cdot)$ terms:
\begin{equation} \label{eq_fregs_loss}
\begin{split}
    \mathcal{L}_{FreGS} = \sum_{s \in \{LP,HP\}} \lambda_{s}[\mathcal{L}_1(|\mathcal{F}_{s}|,|\mathcal{F}^{gt}_{s}|) + \mathcal{L}_1(\angle \mathcal{F}_{s},\angle\mathcal{F}^{gt}_{s})]
\end{split}
\end{equation}
where $\mathcal{F}_{(\cdot)}^{gt}$ and $\mathcal{F}_{(\cdot)}$ denote the GT and render Fourier spectrum, respectively. The subscript $s \in \{LP, HP\}$ indicates low-pass (LP) or high-pass (HP) spectrum, separated from the full spectrum by filtering operations \cite{imageprocessing_book}. $\lambda_{LP}$ and $\lambda_{HP}$ are the  corresponding balancing terms. 

However, evidently, Fourier-based losses like $\mathcal{L}_{FreGS}$ require four terms of significantly different ranges (between LP and HP, then between magnitude and phase). When used alongside the main loss $\mathcal{L}_{3DGS}$, this requires extensive tuning for the weights. Furthermore, it requires defining parameters of the Fourier-space filtering operations that perform LF-HF separation. Those parameters are not detailed in the papers \cite{FreGS, PGDGS}, making the method's implementation ambiguous and subjected to tuning. As seen in Fig.~\ref{fig_arch_dwtgs}, we implement element-wise Gaussian masks for LF-HF separation, but the mask's standard deviation is hardly interpretable in the Fourier space. 

Moreover, $\mathcal{L}_{FreGS}$'s effects under sparse views remain unclear. In the gradient heatmap of $\mathcal{L}_{FreGS}$ in the middle of Fig. 1, the loss is biased towards the HF details around the leaves, even if our implementation sets higher weights for the LF spectrum. This is because $\mathcal{L}_{FreGS}$ and $\mathcal{L}_{3DGS}$ do not struggle in learning the much smoother and less widely-varying LF details. While HF-centric supervision is important in many tasks \cite{WaveletConvolution, WaveletKnowledgeDistill}, it can be detrimental to sparse-view 3DGS, because enforcing HF consistency may exacerbate overfitting to the limited training views, especially to the severely under-constrained HF details. We support this quantitatively in Section~\ref{sec_ablation}.

\subsection{Wavelet-space Joint Spatial-Frequency Regularization} \label{subsec_wavelet_loss}

The above discussions motivate us towards wavelet-space, where joint spatial-frequency regularization is more compatible with sparse-view 3DGS. Firstly, we define a LF-centric subloss:
\begin{equation} \label{eq_dwtgs_loss_lf}
    \mathcal{L}_{DWTGS\_LF} = \sum_{n=1}^{N}\lambda_{LL}\mathcal{L}_1(\mathbf{X}^n_{LL}, \mathbf{X}_{LL}^{n\_{gt}})
\end{equation}
Compared to \eqref{eq_fregs_loss}, we only supervise the LF LL subband at each DWT level $n \leq N$, where $N$ is the highest DWT level. $\lambda_{LL}$ is the corresponding balancing term, which is easier to control than the balancing terms in $\mathcal{L}_{FreGS}$ as the LL subband is in the same space as the original image, only a coarser version. Furthermore, we enforce  sparsity on the solely HF subband, the HH, with another subloss similar to \cite{TriNeRFLet}, and at novel views similar to \cite{SparseGS, HowtouseDiffusion, CoR-GS, SIDGaussian} for increased generalizability:
\begin{equation} \label{eq_dwtgs_loss_hf}
    \mathcal{L}_{DWTGS\_HF} = \mathcal{L}_{1}( \mathbf{X}^1_{HH}, \mathbf{0})
\end{equation}
Compared to fully supervised HF learning as in \eqref{eq_fregs_loss}, this self-supervised subloss aids in reducing HF overfitting. Note that we restrict HF regularization to the 1-level DWT, because higher levels often carry crucial LF information even in the HF subbands, as seen in Fig.~\ref{fig_vis_subband_multi}. Furthermore, the HF are already implicitly regularized by the multi-level, LL-only loss $\mathcal{L}_{DWTGS\_LF}$, as low-level LL subbands still retain a fair amount of HF \cite{WINE}. 

\begin{table}[htbp]
\caption{Quantitative results, 3‑view LLFF \cite{LLFF} and 12‑view Mip‑NeRF 360 \cite{MipNeRF360} datasets}
\centering
\resizebox{\columnwidth}{!}{%
\begin{tabular}{|c|c|c|c|c|c|c|}
\hline
\multirow{2}{*}{Method} &
\multicolumn{3}{c|}{LLFF \cite{LLFF}} & \multicolumn{3}{c|}{Mip‑NeRF 360 \cite{MipNeRF360}} \\ \cline{2-7}
 & PSNR (↑) & SSIM (↑) & LPIPS (↓) & PSNR (↑) & SSIM (↑) & LPIPS (↓) \\ \hline
3DGS \cite{3DGS} & 19.22 & 0.649 & 0.229 & 18.52 & 0.523 & 0.415 \\
DropGaussian \cite{dropgaussian} & 20.39 & 0.706 & 0.198 & 19.30 & 0.564 & 0.352 \\
FreGS \cite{FreGS}     & 20.40 & 0.706 & 0.198 & 19.41 & 0.571 & 0.344 \\
DWTGS & \textbf{20.78} & \textbf{0.720} & \textbf{0.197} & \textbf{19.75} & \textbf{0.590} & \textbf{0.343} \\ \hline
\end{tabular}}
\label{tab_quant_llffmipnerf}
\end{table}

\begin{table}[htbp]
\caption{Quantitative results, 8-view Blender \cite{NeRF} dataset}
\centering
\begin{tabular}{|c|c|c|c|}
\hline
\multirow{2}{*}{Method} & \multicolumn{3}{c|}{Blender \cite{NeRF}} \\ \cline{2-4}
 & PSNR (↑) & SSIM (↑) & LPIPS (↓) \\ \hline
3DGS \cite{3DGS} & 21.56 & 0.847 & 0.130 \\
DropGaussian \cite{dropgaussian} & 25.13 & 0.869 & 0.101 \\
FreGS \cite{FreGS} & 25.20 & 0.870 & 0.103 \\
DWTGS & \textbf{25.56} & \textbf{0.889} & \textbf{0.091} \\ \hline
\end{tabular}
\label{tab_quant_blender}
\end{table}

\begin{table}[htbp]
\caption{Ablation results, 3-view LLFF \cite{LLFF} dataset}
\centering
\begin{tabular}{|c|c|c|c|}
\hline
\multirow{2}{*}{Method} & \multicolumn{3}{c|}{LLFF \cite{LLFF}} \\ \cline{2-4}
 & PSNR (↑) & SSIM (↑) & LPIPS (↓) \\ \hline
DropGaussian \cite{dropgaussian} & 20.39 & 0.706 & 0.198 \\
\hline
+ $\mathcal{L}_{DWTGS\_{LF}}$ (1-level) & 20.50 & 0.711 & 0.214 \\
\rowcolor{green} + $\mathcal{L}_{DWTGS\_{LF}}$ (2-level) & 20.70 & 0.711 & 0.209 \\
+ $\mathcal{L}_{DWTGS\_{LF}}$ (3-level) & 20.64 & 0.711 & 0.208 \\
\hline
\rowcolor{lightred}+ $\mathcal{L}_{DWTGS\_{HF}}$ (sup.) & 20.07 & 0.701 & 0.211 \\
\rowcolor{green} + $\mathcal{L}_{DWTGS\_{HF}}$ (self-sup.) & \textbf{20.78} & \textbf{0.720} & \textbf{0.197} \\
\hline
\end{tabular}
\label{tab_ablation}
\end{table}

\begin{figure}[ht]
    \centering \includegraphics[width=\linewidth]{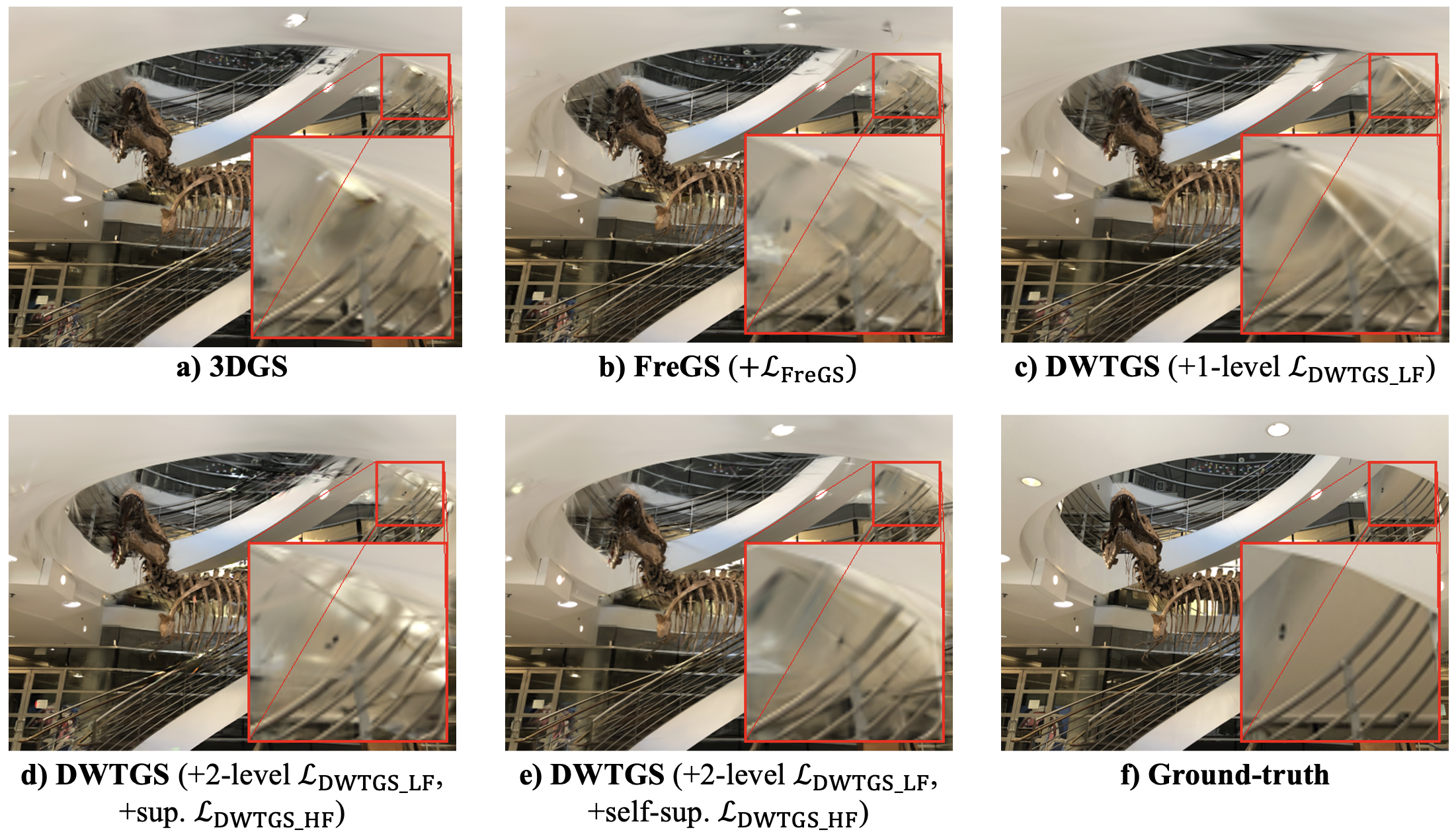}
    \caption{Visual ablations on a novel view region (a) \cite{LLFF}. LF-centric wavelet-space losses at c), d) and e) enforce better structural consistency than Fourier-space loss at b). However, fully supervising HF at d) degrades image quality. Instead, in e), the HF is self-supervised and enforced to be sparse, which yields the best quality.}
    \label{fig_vis_ablation}
\end{figure}

% Our final loss, which replaces $\mathcal{L}_{FreGS}$ and is illustrated in the top part of Fig.~\ref{fig_arch_dwtgs}, is as follows:
% \begin{equation} \label{eq_dwtgs_loss}
%     \mathcal{L}_{DWTGS} = \mathcal{L}_{DWTGS\_LF} + \mathcal{L}_{DWTGS\_HF}
% \end{equation}

\section{Experiments}

\subsection{Datasets \& Implementation Details}

\textbf{Datasets}. DWTGS is evaluated on the LLFF \cite{LLFF} (3  views), Mip-NeRF 360 \cite{MipNeRF360} (12 views) and Blender \cite{NeRF} (8 views) datasets. Novel view synthesis performance of 3DGS is evaluated on held-out test images. We adopt PSNR, SSIM \cite{SSIM} and LPIPS \cite{LPIPS} as evaluation metrics, and mainly compare DWTGS with its Fourier-space predecessor, FreGS \cite{FreGS}.

\textbf{Implementation Details}. Our implementations are built on DropGaussian \cite{dropgaussian}, which is based on the original 3DGS \cite{3DGS}, but with a Gaussian random dropping strategy that boosts sparse-view performances. Due to lack of code availability, we re-implement the Fourier-space losses ourselves. For LF-HF separation, we leverage a LP Gaussian element-wise mask that only retains 50\% of all frequencies, and use its inverse as the HP mask. For progressive HF supervision, we attenuate the HP mask based on the current training iteration, as illustrated by the donut shape in Fig.~\ref{fig_arch_dwtgs}. We only supervise the HF after 5K iterations. The terms $\lambda_{LP}$ and $\lambda_{HP}$ are set to 0.01 to balance with the main objective $\mathcal{L}_{3DGS}$. Those parameters are chosen after extensive experiments to ensure the best performances. In implementing the wavelet-space losses, we set $\lambda_{LL}$ to 0.5 and use $N = 2$ DWT levels. We regularize the HF with \eqref{eq_dwtgs_loss_hf} by rendering novel views every 10 iterations, and stop after 5K iterations to save computation. Finally, we used the Haar wavelet \cite{wavelet-book} for the DWT.

\subsection{Quantitative Results} \label{sec_results}

Tables~\ref{tab_quant_llffmipnerf} and~\ref{tab_quant_blender} show the quantitative results, averaged across all scenes of each dataset. We re-ran all methods on our machine, except 3DGS. Generally, across all datasets, Fourier-space losses provide little performance increases, while our wavelet-space losses increase the PSNR by 0.3-0.4.

\subsection{Ablation Study}
\label{sec_ablation}

We ablate each component of DWTGS in Table~\ref{tab_ablation}. In the 2\textsuperscript{nd} to 4\textsuperscript{th} rows, it is observed that LF-only supervision consistently boosts rendering quality as it mitigates HF hallucinations. Using the higher-level LL subbands further increases performance, due to the stricter LF constraint and implicit regularization of HF details \cite{WINE}, although this effect diminishes with 3-level as the representations become too coarse. In the last two rows, we complemented the 2-level LF supervision with HF regularization. The 5\textsuperscript{th} row complements \eqref{eq_dwtgs_loss_lf} with terms for the HF subbands, in a fully supervised manner. However, this exacerbates overfitting to training views, reducing test performances significantly. On the other hand, the final row leverages the self-supervised, sparsity-enforcing HF loss described at \eqref{eq_dwtgs_loss_hf}, further enhancing generalizability.

\section{Conclusion}

In this paper, we show that sparse-view 3DGS is better regularized using LF signals. Replacing Fourier-space regularization losses that exhibit HF biases, we introduce wavelet-space losses that regularize spatial-frequency jointly. Firstly, we propose supervising multi-level LL subbands of the render and ground-truth images. Secondly, we relegate the HF to self-supervised regularization that enforces sparsity. Evaluations on benchmark datasets show that our LF-centric wavelet losses allow 3DGS to outperform the Fourier baselines non-trivially. 

\textbf{Acknowledgements}. The first author of this work was financially supported by the Vingroup Science and Technology Scholarship Program for Overseas Study for Master’s and Doctoral Degrees.

\bibliographystyle{ieeetr}
\bibliography{references}

\end{document}